\documentclass[10pt,twocolumn,letterpaper]{article}

\usepackage{iccv}
\usepackage{times}
\usepackage{epsfig}
\usepackage{graphicx}
\usepackage{amsmath}
\usepackage{amssymb}

\usepackage{subfigure}
\usepackage{enumitem}
\usepackage{booktabs}
\usepackage{adjustbox}
\usepackage[font=small]{caption}
\newcommand{\ours}{LPD}

\usepackage{hyperref}
\hypersetup{breaklinks=true,colorlinks,linkcolor=blue,citecolor=blue}

\iccvfinalcopy % *** Uncomment this line for the final submission

\def\iccvPaperID{3448} % *** Enter the ICCV Paper ID here

\ificcvfinal\fi

\title{Discovering 3D Parts from Image Collections}

\author{
Chun-Han Yao\\
UC Merced%\\
\and
Wei-Chih Hung\\
Waymo%\\
\and
Varun Jampani\\
Google%\\
\and
Ming-Hsuan Yang\\
UC Merced%\\
}

\begin{document}

%\maketitle
\ificcvfinal\thispagestyle{empty}\fi

\twocolumn[{
\maketitle
\vspace{-3.0em}
\centerline{
\includegraphics[width=\linewidth,trim={4pt 4pt 4pt 4pt}]{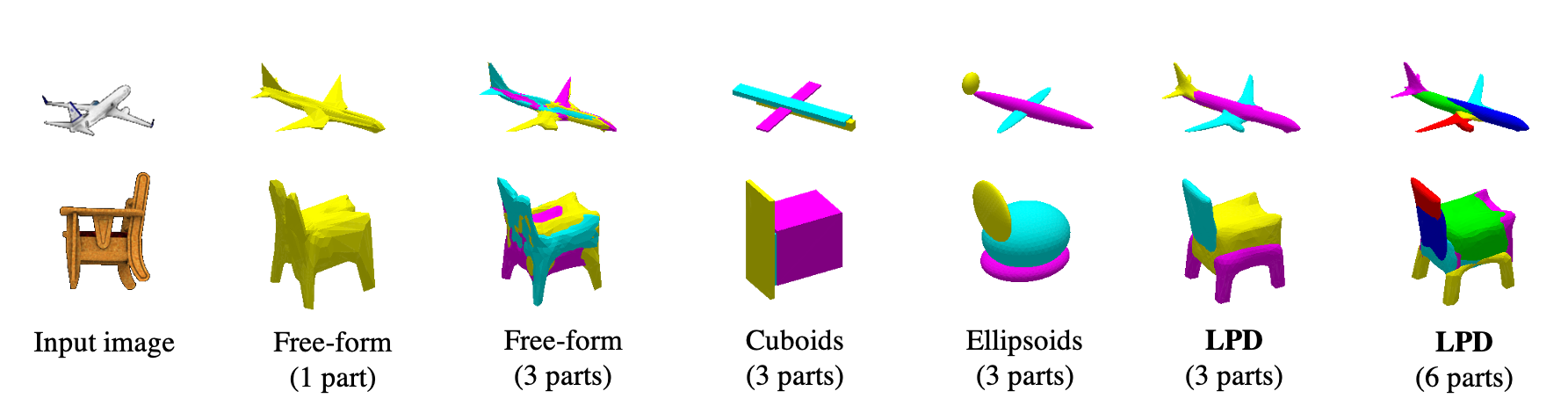}
}
\vspace{-2mm}
\captionof{figure}{\textbf{Discovering 3D parts form single-view image collections.}
Our method (\ours) enables self-supervised 3D part discovery while learning to reconstruct object shapes from single-view images.
Compared to other methods using different part constraints, \ours~discovers more faithful and consistent parts, which improve the reconstruction quality and allow part reasoning/manipulation.}
\label{fig:cover}
\vspace{2em}
}]

%%%%%%%%% ABSTRACT
\begin{abstract}
Reasoning 3D shapes from 2D images is an essential yet challenging task, especially when only single-view images are at our disposal.
While an object can have a complicated shape, individual parts are usually close to geometric primitives and thus are easier to model. 
Furthermore, parts provide a mid-level representation that is robust to appearance variations across objects in a particular category.
In this work, we tackle the problem of 3D part discovery from only 2D image collections.
Instead of relying on manually annotated parts for supervision, we propose a self-supervised approach, latent part discovery (\ours).
Our key insight is to learn a novel part shape prior that allows each part to fit an object shape faithfully while constrained to have simple geometry.
Extensive experiments on the synthetic ShapeNet, PartNet, and real-world Pascal 3D+ datasets show that our method discovers consistent object parts and achieves favorable reconstruction accuracy compared to the existing methods with the same level of supervision.
Our project page with code is at \url{https://chhankyao.github.io/lpd/}.
\end{abstract}

%%%%%%%%% BODY TEXT
\vspace{-3mm}
\section{Introduction}
\vspace{-1mm}
Recognizing and reasoning about objects surrounding us is essential for many computer vision systems.
While deep learning models have been shown to perform well at recognizing~\cite{krizhevsky2012imagenet, simonyan2014very, he2016deep} and localizing~\cite{girshick2015fast, ren2015faster, liu2016ssd} objects in a 2D image, reasoning 3D attributes of objects from a single image remains a challenging task.
Single-view 3D reasoning is fundamentally ill-posed due to several factors that cause ambiguous object appearance in 2D images, \eg, camera pose, self-occlusions, lighting, and material properties.
Although objects in general have complicated shapes, they can usually be decomposed into parts that have simpler geometry and are relatively easy to model.
Furthermore, most object instances of a particular category share similar part configurations, \eg, the wings, body, and tail of airplanes.
In this work, we propose to tackle the problem by discovering faithful and consistent 3D parts from 2D image collections.
Compared to existing single-view 3D reconstruction approaches that directly predict an object shape, we aim to learn rich and dense part configurations which form an entire object when combined.

% Why single-view training
%
Although several recent methods~\cite{tulsiani2017learning, mandikal20183d, chen2019bae, gao2019sdm, li2020learning, luo2020learning, paschalidou2020learning, kawana2020neural} leverage part-based representations for 3D object reasoning, they rely on either 3D object shapes or explicit part annotations as supervision.
Moreover, the learned parts only serve as additional information and are not exploited to improve 3D reconstruction.
Considering that collecting ground-truth 3D shapes and the corresponding part labels is labor intensive, we follow a practical scenario where only single-view images, 2D object silhouettes, and camera viewpoints are available for model training.
In contrast to existing techniques, our method automatically discovers 3D parts from image collections in a self-supervised manner.

% Why learn part prior
%
A common practice to represent 3D parts is to use geometric primitives such as ellipsoids or cuboids~\cite{tulsiani2017learning}.
They provide a strong regularization of part shapes but are usually too coarse to faithfully represent object parts.
%
% We show example reconstruction results using cuboid part primitives in Figure~\ref{fig:cover}. 
%
As an alternative, several approaches adopt meshes~\cite{kato2018neural, wang2018pixel2mesh, gkioxari2019mesh, kato2019learning, liu2019soft, henderson2020leveraging, goel2020shape, li2020self} or point-cloud~\cite{fan2017point, insafutdinov2018unsupervised, l2019differ, navaneet2019capnet, deprelle2019learning} representations.
Although these representations are more expressive and can faithfully describe part shapes, they lack part shape regularization that is particularly needed in a weakly-supervised or unsupervised setting.
The key insight of this work is to represent 3D parts with deep latent embeddings.
Specifically, we propose to learn a prior distribution of part shapes with a variational auto-encoder (VAE)~\cite{kingma2013auto} that encodes part shapes as latent embeddings.
We call this network \textit{Part-VAE} and pre-train it with a set of geometric primitives like cones, cylinders, cuboids, and ellipsoids.
%
%Our key observation is that part encodings learned using simple geometric primitives are expressive enough to represent object parts while also being regularized to be close to the primitive shapes.
%
We then learn a reconstruction network that takes an input image and predicts part embeddings to obtain a 3D mesh by passing through the decoder of Part-VAE.
%
%We combine these predicted part meshes to obtain complete object mesh.
%
%We train part-VAE along with the reconstruction network using both primitive shapes and given image collections.
%
To further improve the quality of part discovery and reconstruction, we propose a novel part adversarial loss which involves re-assembling parts from different objects in the same category.
We name the proposed method {\it latent part discovery (\ours)}.
Figure~\ref{fig:cover} shows several reconstruction results with and without part prior, which demonstrate that \ours~can discover consistent parts and produce faithful reconstruction to the input image.

We evaluate \ours~on the synthetic ShapeNet~\cite{chang2015shapenet}, PartNet~\cite{mo2019partnet} and the real-world Pascal 3D+~\cite{xiang2014beyond} datasets. 
Both quantitative and qualitative results demonstrate that our approach achieves favorable performance against the state-of-the-art methods using the same level of supervision.
In addition to part discovery, our part representation enables object manipulation like selective part swapping, interpolation, and random shape generation from the latent space.
In this work, we make the following contributions:
\vspace{-1mm}
\begin{itemize}[leftmargin=*]
    \item We propose a part-based single-view 3D reasoning network which can automatically discover object parts.
    To the best of our knowledge, this is the first work that discovers 3D parts in a self-supervised manner without using any 3D shape or multi-view supervision.
    \vspace{-1mm}
    \item We develop Part-VAE to learn a latent prior over part shapes.
    We show that training with geometric primitives can learn useful part embeddings, allowing each part to faithfully represent object shape while constrained to have simple geometry.
    \vspace{-1mm}
    \item We conduct extensive experiments on both synthetic and natural images.
    % to demonstrate that our approach achieves favorable reconstruction accuracy against the state-of-the-art techniques.
    %
    Qualitatively, our method produces more faithful and consistent object parts compared to other part-based methods.
    Quantitatively, the discovered parts improve whole-object reconstruction and achieve favorable accuracy against the state-of-the-art techniques.
    In addition, our Part-VAE allows us to manipulate object parts for various applications.
\end{itemize}
\vspace{-2mm}

\vspace{-1mm}
\section{Related Work}
\vspace{-1mm}
\noindent \textbf{3D Reconstruction.}
While 3D representation has been widely studied for decades, the best and unified way to represent general objects remains unclear.
%
% In this section, we categorize the existing 3D reconstruction methods by their shape representation.
%
Voxel grids~\cite{choy20163d, tulsiani2017multi, tulsiani2018multi, li2019synthesizing}, 
point clouds~\cite{fan2017point, insafutdinov2018unsupervised, l2019differ, navaneet2019capnet, deprelle2019learning}, 
and meshes~\cite{kato2018neural, wang2018pixel2mesh, gkioxari2019mesh, kato2019learning, liu2019soft, chen2019learning, henderson2020leveraging, goel2020shape, li2020self} 
are commonly used to represent object shapes.
Several recent methods~\cite{genova2019learning, mescheder2019occupancy, yamashita20193d, hao2020dualsdf, deng2020cvxnet, genova2020local} explore the possibilities to represent 3D shapes in a functional space.
%
% For instance, Mescheder~\etal~\cite{mescheder2019occupancy} propose OccNet which predicts whether a given 3D point is inside an object.
%
% Genova~\etal~\cite{genova2019learning} learn object templates with a set of structured implicit functions (SIF).
%
% To have more regularized local regions, CvxNet~\cite{deng2020cvxnet} and GMNet~\cite{yamashita20193d} represent object shapes with convex functions and Gaussian mixtures, respectively.
%
While fine-grained voxels, point clouds, and local functions can represent complicated shapes, the flexibility of representation demands strong 3D or multi-view supervision for training.
%
% In addition, these representations may entail additional computational cost to reconstruct object surfaces that are amenable for rendering.
%
Meshes, on the other hand, are constrained to form a water-tight surface and are convenient to render 2D images.
By deforming from a simple template mesh like sphere or cuboid, it is easier to apply shape regularization and thus can be applied to reconstruction scenarios with weaker supervision.
One can learn single-view mesh reconstruction from multi-view or single-view images using naive 2D projection or differentiable rendering~\cite{kato2018neural, liu2019soft, chen2019learning}.
For instance, Henderson~\etal~\cite{henderson2020leveraging} generate background images and object rendering to learn textured mesh reconstructions from natural images.
Kato~\etal~\cite{kato2019learning} propose view prior learning (VPL) to improve the reconstructed shapes from unseen views.
Although they are effective for compact and deformable objects, a single mesh cannot represent complicated shapes with holes or disconnected parts.
In this work, we exploit multi-mesh part representation which allows disconnected parts in an object while each part can be well-regularized.

\noindent \textbf{Part Discovery.}
Parts provide a mid-level representation that is robust to appearance variations across objects in the same category.
%
% Parts are shown to be useful for training settings where the supervisory signal for a whole object is weak or unavailable.
%
% For instance, 
Hung~\etal~\cite{hung2019scops} learn 2D co-part segmentation on image collections with self-supervision.
Lathuili{\`e}re~\etal~\cite{lathuiliere2020motion} exploit motion cues in videos for part discovery.
In the 3D domain, Tulsiani~\etal~\cite{tulsiani2017learning} use volumetric cuboids as part abstractions to learn 3D reconstruction.
Li~\etal~\cite{li2020learning} assume known part shapes and learn to assemble them given an input image.
Using a point-cloud representation,
Mandikal~\etal~\cite{mandikal20183d} predict part-segmented 3D reconstructions from a single image and Luo~\etal~\cite{luo2020learning} learn to form object parts by clustering 3D points.
Paschalidou~\etal~\cite{paschalidou2020learning} propose hierarchical part decomposition (HPD) by constraining 3D points with super-quadratic functions~\cite{paschalidou2019superquadrics}.
These methods require 3D ground-truth shape of a whole object or its parts as supervision.
Furthermore, the part shapes in \cite{tulsiani2017learning, paschalidou2020learning} are limited by the expressiveness of their representations.
Li~\etal~\cite{li2020self} leverage 2D semantic parts to improve single-view 3D reconstruction, which, however, does not produce individual part shapes.
To the best of our knowledge, we propose the first 3D part reconstruction method without any part annotations or ground-truth 3D shapes for training, and our latent part representation enables each part to fit a given object shape faithfully.

\begin{figure*}[t]
    \begin{center}
    \includegraphics[width=0.99\linewidth]{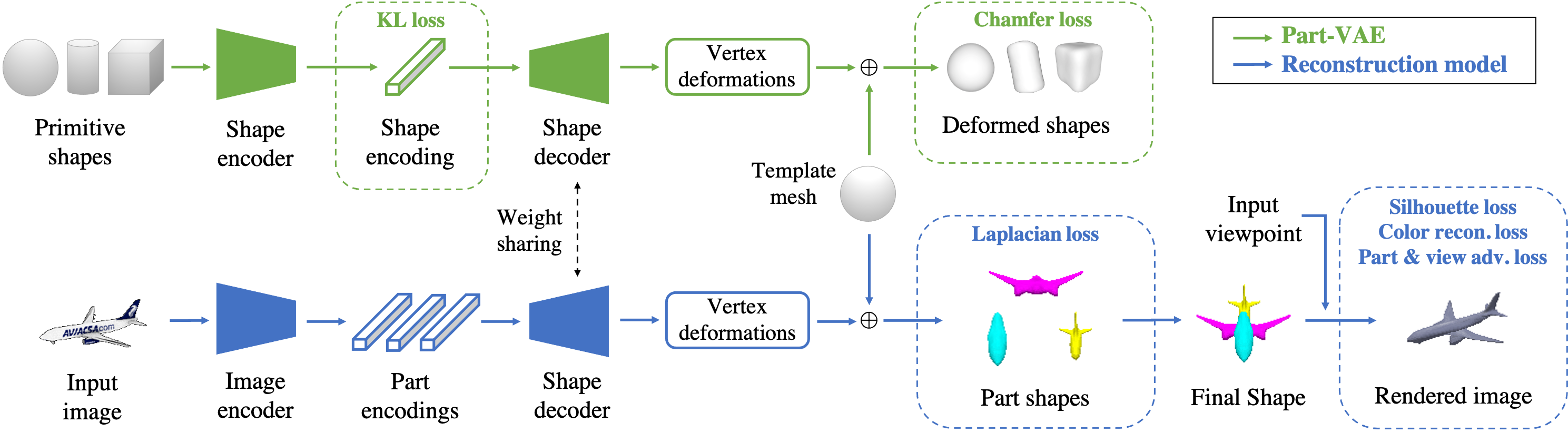}
    \end{center}
    \vspace{-5mm}
    \caption{\textbf{Approach overview.} 
    (top) Our Part-VAE is trained with geometric primitives.
    (bottom) Our reconstruction model shares the shape decoder with Part-VAE and predicts object parts.
    We then composite the reconstructed parts to form a 3D object.
    The prediction of part centroids and surface texture are detailed in the supplemental material.
    % %
    % In the upper half we show the training process of our Part-VAE.
    % %
    % We exploit the pre-trained shape decoder from Part-VAE and train the reconstruction model, as shown in the lower half.
    }
\label{fig:recon}
\end{figure*}

\vspace{-1mm}
\section{Approach}
\vspace{-1mm}
Reasoning 3D objects from single-view images is inherently ill-posed since the reconstructed object can over-fit to a given view and be highly deformed in the unseen parts.
%
% To address this, we propose primitive part prior to regularize 3D shapes in the latent space, and apply part adversarial learning to train a single-view reconstruction model.
%
To address this, we propose \ours~to represent an object with multiple latent parts.
Our intuition is that a complicated object shape can be expressed by assembling simple and regularized parts.
%
% In this section, we introduce the proposed method for discovering latent parts considering the task of single-view 3D reconstruction. %We name the proposed method part variational autoencoder (part-VAE).
%
Consistent with some recent approaches~\cite{kato2019learning, henderson2020leveraging}, we propose a method under a weakly-supervised setting where only a single-view image, 2D object silhouette, and its camera viewpoint are available for each object.
That is, we do not assume any 3D shape or multi-view images to supervise part discovery or reconstruction.
In order to automatically discover underlying parts while reconstructing an object, we propose to learn part embeddings with a variational auto-encoder~\cite{kingma2013auto} (VAE) named \textit{Part-VAE}.
We train a reconstruction model that predicts part embeddings which are then decoded into part meshes to compose the whole object.
Figure~\ref{fig:recon} illustrates the proposed method with two main modules: Part-VAE and reconstruction network.

\vspace{-1mm}
\subsection{Learning Part Prior with Part-VAE}
\label{sec:part_prior}
\vspace{-1mm}
We propose Part-VAE to learn a latent shape prior for object parts. 
The proposed method constrains parts with primitive shapes while allowing the flexibility to fit real-world object parts.
In addition, it enables smooth part-interpolation and novel shape generation by random sampling in the latent space. 
Figure~\ref{fig:recon}~(top) illustrates the training process of the Part-VAE with geometric primitives.
We first collect a set of primitive shapes such as ellipsoids, cylinders, cones, and cuboids, which are centered at origin but with random scaling and rotation.
The Part-VAE network consists of a shape encoder and a shape decoder.
The encoder transforms each given primitive shape to a low-dimensional shape encoding, and the decoder reconstructs the input shape by predicting the vertex deformations of a spherical template mesh.
%
% Since the network output, \ie vertex deformations, has a fixed dimension, we upsample the primitive meshes to have the same number of vertices as the template mesh.
%
To supervise the Part-VAE pre-training, we calculate the Chamfer distance between the input and output vertices as the loss function since the point sets are unordered and not densely corresponding.
Given the vertices of an input shape $Q$ and its reconstruction $P$, the Chamfer loss $\mathcal{L}_{c}$ can be expressed as:
\vspace{-1mm}
%\begin{multline}
%    \mathcal{L}_{c}(P, Q) = 
%    \frac{1}{\| P \|} \sum_{p \in P} 
%    \hspace{1mm} \min_{q \in Q}
%    \big\| p - q \big\|_2^2 
%    \hspace{1mm} + \\
%    \frac{1}{\| Q \|} \sum_{q \in Q} 
%    \hspace{1mm} \min_{p \in P}
%    \big\| p - q \big\|_2^2 .
%\label{eqn:chamfer}
%\end{multline}

\begin{equation}
\begin{split}
    \mathcal{L}_{c}(P, Q) = & 
    \frac{1}{\| P \|} \sum_{p \in P} 
    \hspace{1mm} \min_{q \in Q}
    \big\| p - q \big\|_2^2 
    \hspace{1mm} + \\
    & \frac{1}{\| Q \|} \sum_{q \in Q} 
    \hspace{1mm} \min_{p \in P}
    \big\| p - q \big\|_2^2 .
\label{eqn:chamfer}
\end{split}
\end{equation}
To encourage the continuity of latent shape distribution, we adopt a standard KL divergence loss $\mathcal{L}_{kl}$ calculated between the shape embeddings and a standard normal distribution $\mathcal{N}\sim(0, 1)$.
%
% We further apply Laplacian regularization on the reconstructed mesh vertices to produce smooth surfaces.
%
% We adopt the standard Auto-encoding Variational Bayes~\cite{kingma2013auto} as the VAE model.
%
The overall training loss of Part-VAE is:
$\mathcal{L}_{vae} = 
\mathcal{L}_{c} + 
\lambda_{kl} \hspace{1mm} \mathcal{L}_{kl}$
, where $\lambda_{kl}$ is a weight parameter.
%
% The readers are encouraged to refer to the original VAE paper for more details.
%
% An illustration of the proposed Part-VAE is shown in the upper half of Figure~\ref{fig:recon}.
%
\begin{figure*}[t]
    \begin{center}
    \includegraphics[width=0.99\linewidth]{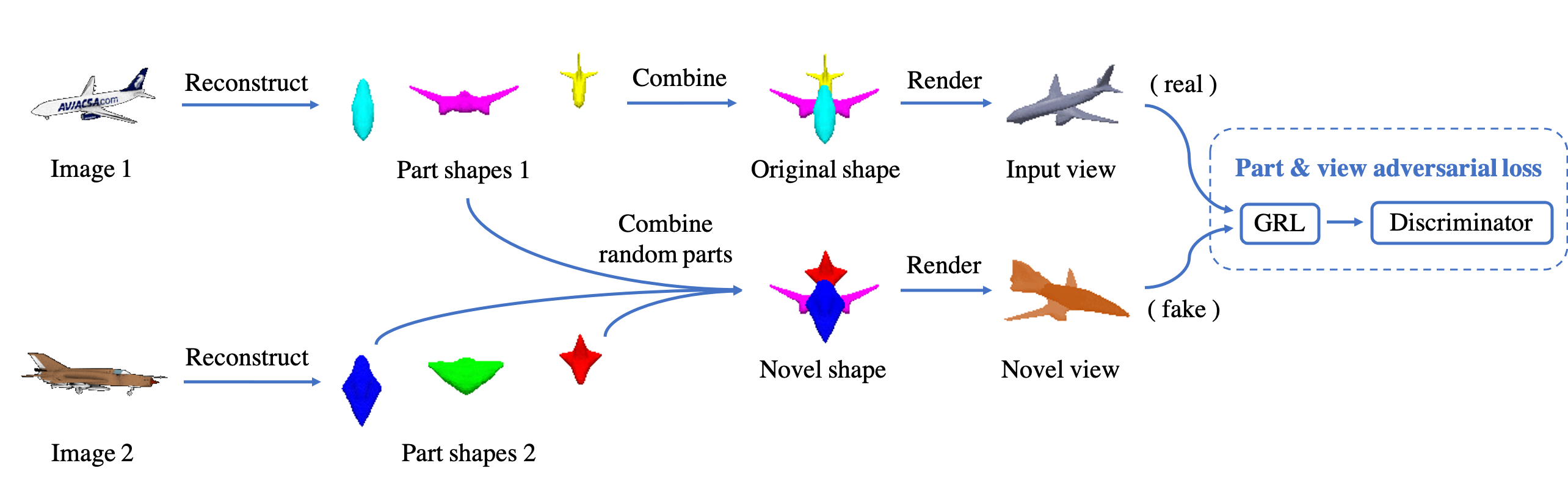}
    \end{center}
    \vspace{-5mm}
    \caption{\textbf{Part and view adversarial learning.}
    % An illustration of the proposed part and view adversarial learning.
    % %
    Given two images with different objects, we randomly combine their reconstructed parts into a novel shape.
    The novel shape is then rendered from a novel viewpoint, which we treat as a `fake' sample.
    We train a discriminator to distinguish the fake and real rendered images.
    By using a gradient reversal layer (GRL), the reconstruction model learns to produce parts that can compose realistic novel shapes. 
    }
\label{fig:adversarial}
\end{figure*}

\vspace{-1mm}
\subsection{Part Discovery by Learning to Reconstruct}
\label{sec:recon}
\vspace{-1mm}
Figure~\ref{fig:recon} (bottom) illustrates our reconstruction model.
Instead of directly predicting an object mesh, we learn an image encoder that takes an input image and predicts the 3D part centroid, latent shape encodings, and surface texture for each part.
The part encodings are then passed through the shape decoder of Part-VAE to generate part meshes.
To compose an entire object, we simply shift the mesh vertices using the predicted part centroids and concatenate the vertices and surfaces of each part.
In the remainder of the paper, we denote the reconstruction model as $R(\cdot)$, which takes an image as input and outputs a part-composited mesh.
The rendering function from viewpoint $v$ is denoted as $G(\cdot, v)$, which produces a rendered image of the input mesh.
Note that each training image $I$ includes a silhouette channel $I_s$ and RGB color channels $I_c$.
Likewise, the rendering function $G$ can be separated into $G_s$ and $G_c$ for silhouette projection and color rendering, respectively.

\vspace{1mm}
\noindent \textbf{Shape Reconstruction Losses.}
To supervise shape reconstruction, we enforce the 2D projection of a reconstructed shape to be close to the ground-truth silhouette.
In particular, we render the predicted meshes with input viewpoints using a differentiable renderer~\cite{liu2019soft}, and calculate the intersection-over-union (IoU) ratio between the rendered and ground-truth silhouettes.
Then the silhouette loss $\mathcal{L}_{sil}$ is computed as:
\begin{equation}
    \mathcal{L}_{sil}(I, v) = \frac
    {\lVert I_s \odot G_s(R(I), v) \rVert_{1}}
    {\lVert I_s + G_s(R(I), v) \rVert_{1} - 
     \lVert I_s \odot G_s(R(I), v) \rVert_{1}} ,
\label{eqn:silhouette}
\end{equation}
%
%where $I_s$ is the ground-truth silhouette of the input %image $I$, $R(.)$ is the reconstruction network, $G_s(., %v)$ is the silhouette rendering with viewpoint $v$, and 
%
where $\odot$ denotes element-wise multiplication.
We further apply Laplacian regularization $\mathcal{L}_{lap}$ on the reconstructed mesh vertices:
%
% The Laplacian loss  is defined as:
%
\vspace{-1mm}
\begin{equation}
    \mathcal{L}_{lap}(P) = 
    \frac{1}{\| P \|} \sum_{p \in P}
    \Big\| \hspace{1mm} p - \frac{1}{\| \mathcal{N}(p) \|} 
    \sum_{q \in \mathcal{N}(p) } q  \hspace{1mm} \Big\|_2^2 ,
\label{eqn:laplacian}
\end{equation}
where $\mathcal{N}(p)$ denotes the neighboring vertices of vertex $p$.
It aims to smooth the mesh surfaces by pulling each vertex towards the center of its neighboring pixels.
Note that this regularization is applied on each part mesh individually so discontinuity between part surfaces is allowed.

\vspace{1mm}
\noindent \textbf{Color Reconstruction Loss.}
We further exploit the color information in input images by generating textured reconstructions.
Given the part encodings, our model predicts texture flow to map the input image to a UV texture image.
We then color the mesh surfaces by sampling from the texture image using a pre-defined UV mapping function.
The texture flow is predicted per object part so each part has more coherent texture.
We denote the overall color rendering process as $G_c$ and show more details in the supplemental material.
The color reconstruction loss $\mathcal{L}_{cr}$ is defined on the semantic features of input and rendered images:
\begin{equation}
    \mathcal{L}_{cr}(I, v) = 
    {\lVert F(I_c) - F(G_c(R(I), v)) \rVert_{2}^2} ,
\label{eqn:color}
\end{equation}
where $F$ is the feature extractor of a fixed classification network.
We use AlexNet~\cite{krizhevsky2012imagenet} pre-trained on the ImageNet dataset~\cite{krizhevsky2012imagenet} and extract the output of multiple convolutional layers as $F$.
It encourages the rendered image to be perceptually similar to the input at different levels.

\begin{figure*}[ht!]
\vspace{-5mm}
\centering
\setlength\tabcolsep{6pt}
\renewcommand{\arraystretch}{0}
\begin{tabular}[t]{ccccccc}
\includegraphics[width = .12\linewidth]{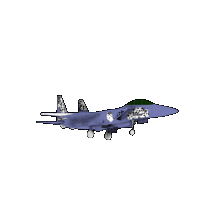} & 
\includegraphics[width = .12\linewidth]{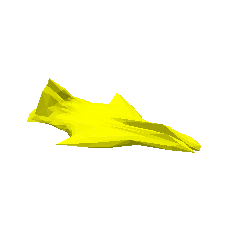} & \includegraphics[width = .12\linewidth]{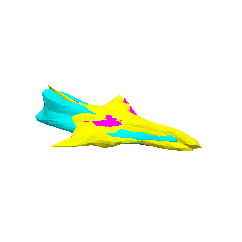} &
\includegraphics[width = .12\linewidth]{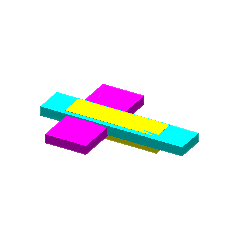} &
\includegraphics[width = .12\linewidth]{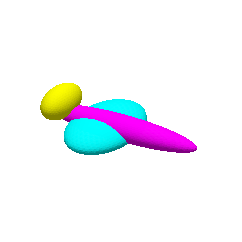} &
\includegraphics[width = .12\linewidth]{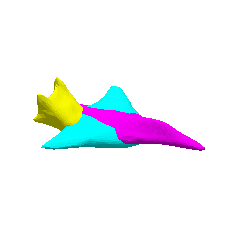} &
\includegraphics[width = .12\linewidth]{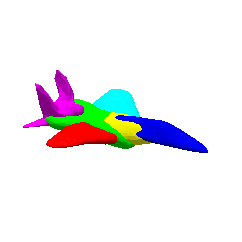}
\vspace{-10mm}
\\
\includegraphics[width = .12\linewidth]{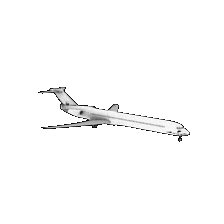} & 
\includegraphics[width = .12\linewidth]{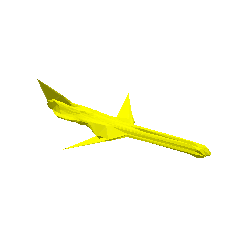} & \includegraphics[width = .12\linewidth]{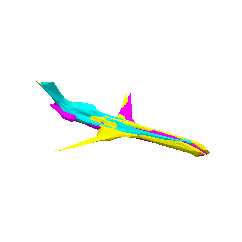} &
\includegraphics[width = .12\linewidth]{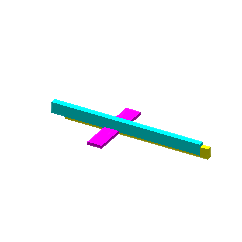} &
\includegraphics[width = .12\linewidth]{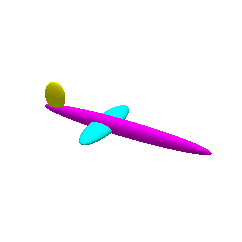} &
\includegraphics[width = .12\linewidth]{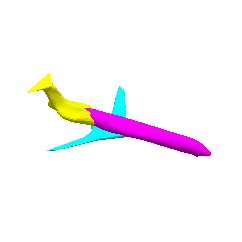} &
\includegraphics[width = .12\linewidth]{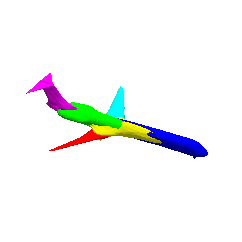}
\vspace{-8mm}
\\
\includegraphics[width = .12\linewidth]{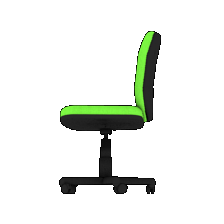} & 
\includegraphics[width = .12\linewidth]{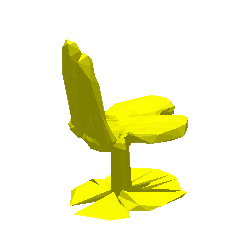} & \includegraphics[width = .12\linewidth]{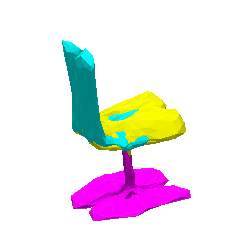} &
\includegraphics[width = .12\linewidth]{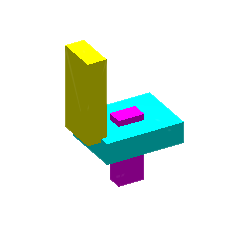} &
\includegraphics[width = .12\linewidth]{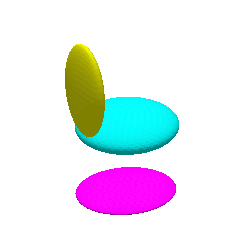} &
\includegraphics[width = .12\linewidth]{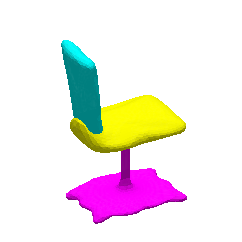} &
\includegraphics[width = .12\linewidth]{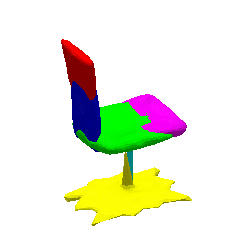}
\vspace{-1mm}
\\
\includegraphics[width = .12\linewidth]{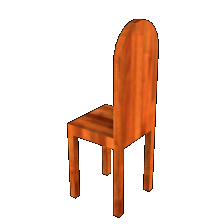} & 
\includegraphics[width = .12\linewidth]{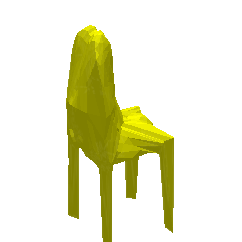} & \includegraphics[width = .12\linewidth]{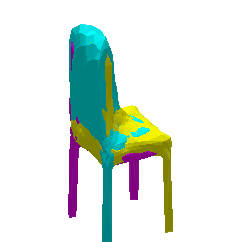} &
\includegraphics[width = .12\linewidth]{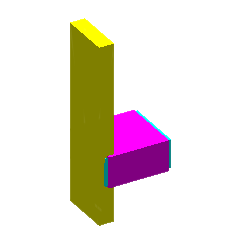} &
\includegraphics[width = .12\linewidth]{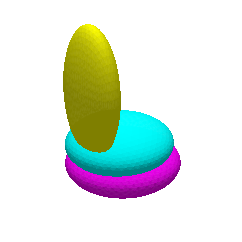} &
\includegraphics[width = .12\linewidth]{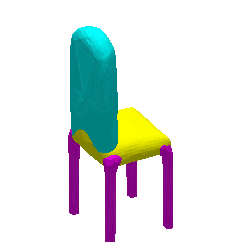} &
\includegraphics[width = .12\linewidth]{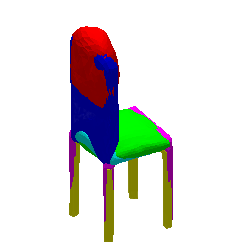}
\vspace{2mm}
\\
Input & 
VPL~\cite{kato2019learning} & 
Free-form & 
Cuboids & 
Ellipsoids &
\ours & 
\ours
\\
 & (1 part) & (3 parts) & 
(3 parts) & (3 parts) & (3 parts) & (6 parts)
\end{tabular}
\vspace{-1mm}
\caption{
\textbf{Qualitative results on the ShapeNet dataset~\cite{chang2015shapenet}.}
\ours~models (ours) adopt the proposed Part-VAE and adversarial learning.
The 3-part free-form model reconstructs a whole object with three fully-deformable meshes without any part prior.
Compared to the baselines, our approach can produce more faithful and consistent parts from diverse objects.
}
\label{fig:shapenet}
\end{figure*}

\vspace{-1mm}
\subsection{Part and View Adversarial Learning}
\vspace{-1mm}
Unlike the multi-view or 3D-supervised settings, single-view training requires stronger regularization to produce realistic 3D shapes and to discover meaningful parts.
Based on the intuition that object parts are interchangeable and they should look realistic from various viewpoints, we extend the view adversarial learning in VPL~\cite{kato2019learning} with part adversarial learning.
As illustrated in Figure~\ref{fig:adversarial}, we assemble a novel shape by randomly combining parts from different objects of the same class in a training batch, then render the novel shape from a novel viewpoint.
To make the novel shape realistic, we treat the rendered images of novel shapes as fake examples and those of original shapes as real ones.
A discriminator is then trained to classify each rendered image as real or fake.
%
% These rendered images are then used as negative (fake) samples to train a discriminator which aims to classify the real and fake images.
%
% Following standard adversarial training, the adversarial loss $\mathcal{L}_{adv}$ is applied to fool the discriminator:
%
We train the discriminator with a binary cross-entropy between the positive and negative samples:
%
%\VJ{$G$ in this equation is not defined. Is it $G_c$? Also, please make the figure consistent accordingly. This equation and explanation can be improved. There are two images in figure-3, whereas $R'(I)$ only takes single image as input.}
%
%\begin{multline}
%    \mathcal{L}_{adv} (I, I', v) = 
%    - \text{log} (D(G(R(I), v))) \\
%    - \text{log} %(1-D(G(R'(I, I'), v'))) ,
%\label{eqn:adversarial}
%\end{multline}
\begin{equation}
\begin{split}
    \mathcal{L}_{adv} (I, I', v)  = 
    & - \text{log} (D(G(R(I), v))) \\
    & - \text{log} (1-D(G(R'(I, I'), v'))) ,
\label{eqn:adversarial}
\end{split}
\end{equation}
where $I'$ is a random image different from the input $I$, $R'(\cdot, \cdot)$ is the reconstruction model with random part selection from two input images, $v'$ is a random novel view, and $D$ is the discriminator.
To apply adversarial training, we add a gradient reversal layer (GRL)~\cite{ganin2014unsupervised} before the discriminator.
As a result, the reconstruction model is trained to fool the discriminator by generating novel objects with realistic shapes.
Considering that different object classes may have different view and shape prior, we condition the discriminator with input class labels during training.
Note that the class labels are not required during inference.
That is, for given data, we train a single part-discovery/reconstruction model that operates across different object categories.
This part-based adversarial learning approach is proposed as a semantic constraint to make the global part arrangements more feasible and realistic.

\begin{table*}[ht!]
\centering
\caption{\textbf{Ablative evaluations on the ShapeNet dataset~\cite{chang2015shapenet}.}
The base model reconstructs an object shape with 3 meshes, each is fully-deformable as in SoftRas~\cite{liu2019soft}
(PP: part prior, VA: view adversarial learning, PA: part adversarial learning, CR: color reconstruction).}
\vspace{-2mm}
\small
\setlength\tabcolsep{3.6pt}
\begin{tabular}{cccc ccccc ccccc cccc}
\toprule
    PP &
    VA &
    PA &
    CR &
    Airplane &
    Bench &
    Dresser &
    Car &
    Chair &
    Display &
    Lamp &
    Speaker &
    Rifle &
    Sofa &
    Table &
    Phone &
    Vessel &
    All 
\\
\midrule
     & \checkmark & \checkmark & \checkmark &
     56.7 & 34.4 & 56.0 & 68.3 & 43.1 &
     34.8 & 47.2 & 59.9 & 50.6 & 48.5 &
     41.1 & 42.7 & 53.4 & 49.0
\\
    \checkmark & & \checkmark & \checkmark &
    57.2 & 36.2 & 60.7 & 72.2 & 44.1 &
    39.8 & 48.2 & 63.6 & 51.9 & 49.6 &
    43.3 & 51.5 & 55.1 & 51.8
\\
    \checkmark & \checkmark &  & \checkmark &
    57.1 & 35.8 & \bf{61.4} & 73.7 & 45.1 & 
    39.4 & 48.5 & 63.7 & 52.7 & 49.3 & 
    43.9 & 52.5 & 54.9 & 52.2
\\
    \checkmark & \checkmark & \checkmark & &
    57.1 & 36.0 & 61.0 & 74.1 & 45.2 & 
    39.7 & 48.5 & \bf{63.8} & 53.0 & 49.7 & 
    43.9 & 52.2 & 55.1 & 52.3
\\
    \checkmark & \checkmark & \checkmark & \checkmark &
    \bf{57.3} & \bf{37.3} & 60.9 & \bf{75.2} & \bf{45.5} &
    \bf{40.8} & \bf{49.6} & 63.3 & \bf{54.5} & \bf{50.1} &
    \bf{44.3} & \bf{52.7} & \bf{56.2} & \bf{52.9}
\\
\bottomrule
\end{tabular}
\label{tab:shapenet}
\end{table*}

\vspace{-1mm}
\subsection{Model Training and Inference}
\vspace{-1mm}
We first pre-train the Part-VAE with primitive shapes to minimize the loss $\mathcal{L}_{vae}$.
Next, the Part-VAE and reconstruction network are jointly trained using image collections together with primitive shapes.
The overall objective function of reconstruction network is given by:
\begin{equation}
 %   \min 
    \quad
    \mathcal{L}_{sil} + 
    \lambda_{lap} \hspace{1mm} \mathcal{L}_{lap} + 
    \lambda_{cr} \hspace{1mm} \mathcal{L}_{cr} - 
    \lambda_{adv} \hspace{1mm} \mathcal{L}_{adv} ,
\label{eqn:objective}
\end{equation}
where ($\lambda_{lap}$, $\lambda_{cr}$, $\lambda_{adv}$) are weight parameters.
The discriminator is trained to minimize $\mathcal{L}_{adv}$, whose gradients are reversed and back-propagated to the reconstruction network to perform adversarial learning.
Note that we still fine-tune the Part-VAE with primitive shapes in this stage so that the shape decoder is regularized while adapting to various part shapes in the training images.
The Part-VAE, reconstruction network, and discriminator are parametrized as deep neural networks, and the weights are optimized by mini-batch gradient descent.
In the model inference phase, we discard the Part-VAE encoder and discriminator.
An input image is simply passed through the image encoder and Part-VAE decoder to reconstruct the 3D parts.
We represent each object part by a deformable mesh with $N_v=642$ vertices and $N_f=1280$ faces.
The size of texture image is 64$\times$64.
We implement the proposed method in PyTorch~\cite{paszke2019pytorch} framework and use Adam optimizer~\cite{kingma2014adam} for training.
The hyper-parameters are tuned on a validation set.
We present more model details in the supplemental material, and the source code will be made available to the public.

\vspace{-1mm}
\section{Experiments and Analysis}
\vspace{-1mm}

\noindent \textbf{Metrics and Baselines.}
Evaluating self-supervised part discovery can be ambiguous and subjective as the discovered parts need not correspond to human-annotated ones.
Due to the lack of standard metrics or benchmark for 3D part discovery, we qualitatively compare the discovered parts with other methods.
As a reference, we also quantitatively evaluate the reconstruction accuracy at both object level and part level.
We convert each predicted mesh into a volume of $32^3$ voxels and calculate the intersection-over-union (IoU) ratio between the voxelized object and the ground-truth voxels.
We report results of our model with $k=3$ parts and latent part embedding dimension of $d=64$ if not specified otherwise.
Since our work is the first to discover 3D parts using single-view supervision, we mainly compare \ours~against three part-based baselines: cuboids, ellipsoids, and free-form meshes.
We implement the cuboid and ellipsoid models by reconstructing each object part with a scalable cuboid/ellipsoid.
The free-form model adopts fully-deformable meshes without any part shape prior.
For object-level reconstruction, we evaluate our method against SoftRas~\cite{liu2019soft} and VPL~\cite{kato2019learning} as they adopt a similar training setting with single-view images and known viewpoints.
While there are several other methods for single-view 3D reconstruction, we omit the comparison with them since whole-object reconstruction is not the major focus of this work.
We experiment on the synthetic ShapeNet~\cite{chang2015shapenet}, PartNet~\cite{mo2019partnet}, and real-world Pascal 3D+~\cite{xiang2014beyond} datasets.
%
% We present the main findings here and additional results in the supplemental material.

\begin{figure}[ht!]
\centering
\setlength\tabcolsep{0pt}
\begin{tabular}[t]{ccccc}
\includegraphics[width = .2\linewidth]{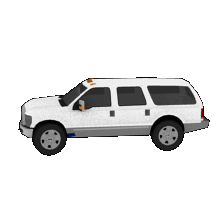} & 
\includegraphics[width = .2\linewidth]{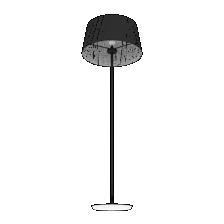} & 
\includegraphics[width = .2\linewidth]{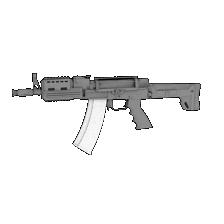} & 
\includegraphics[width = .2\linewidth]{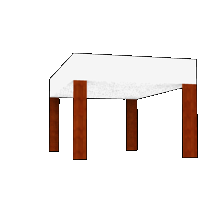} & 
\includegraphics[width = .2\linewidth]{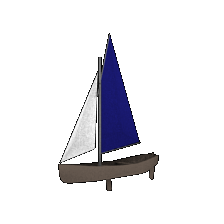}
\\
\includegraphics[width = .2\linewidth]{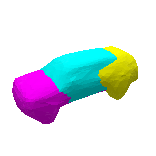} &
\includegraphics[width = .2\linewidth]{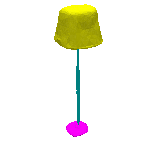} & 
\includegraphics[width = .2\linewidth]{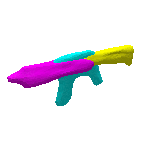} & 
\includegraphics[width = .2\linewidth]{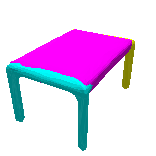} &
\includegraphics[width = .2\linewidth]{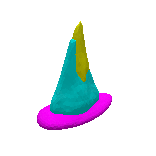}
\vspace{1mm}
\\
Car & Lamp & Rifle & Table & Watercraft
\end{tabular}
\vspace{-1mm}
\caption{\textbf{Generalization across classes.}
We show sample inputs (top) and \ours~results (bottom) of different ShapetNet classes.}
\label{fig:class}
\vspace{-1mm}
\end{figure}

\vspace{-1mm}
\subsection{Results on ShapeNet}
\vspace{-1mm}
We first conduct experiments with the synthetic dataset provided by Kar~\etal~\cite{kar2017learning}, which contains 43,784 objects in 13 classes from ShapeNet~\cite{chang2015shapenet}.
Each sample includes a 3D CAD model, 20 camera viewpoints for rendering, and the corresponding rendered images at a resolution of 224$\times$224 pixels.
We use the same training/validation/testing splits as the original dataset.
The training images are augmented by random shuffling the RGB channels and horizontal flipping.
We only use one view per object for training and evaluate on all the 20 views in the test set (with independent single-view reconstruction on each view).
% 20 views for testing.
%
The ground-truth 3D shapes are only used for testing.
We show some qualitative results of our method and other baselines in Figure~\ref{fig:shapenet}.
More part reconstruction results on car, lamp, rifle, table, and watercraft samples are shown in Figure~\ref{fig:class} to demonstrate that our method generalizes well across diverse object classes.

\vspace{1mm}
\noindent \textbf{Ablations on the Proposed Models.}
We perform ablation studies on the proposed method by removing each component at a time.
As shown in Table~\ref{tab:shapenet}, removing part prior learning causes a significant drop (3.9\%) on the overall reconstruction accuracy.
%
% Table~\ref{tab:shapenet} shows quantitative evaluations of different components in the proposed method.
%
% Compared to the baseline model where each of the three parts deforms freely, the part prior learning provides a significant gain (2.9\%) on the reconstruction accuracy.
%
It shows that the part prior provided by Part-VAE effectively improves the generalization of the discovered parts.
%
% With the use of adversarial learning and color reconstruction, we further boost the voxel IoU by a clear margin (1.9\%).
%
Without the use of adversarial learning and color reconstruction, we also observe lower voxel IoU by a clear margin (0.6-1.1\%).

\begin{table}[t]
\setlength\tabcolsep{4pt}
\small
    \centering
    \caption{\textbf{Voxel IoU results on the ShapeNet dataset~\cite{chang2015shapenet}}. We compare our method with the state-of-the-art single-view supervised and 3D supervised approaches.}
    \vspace{-2mm}
    \begin{tabular}{ll cccc}
    \toprule
        Method & Supervision & 
        Airplane &
        Car &
        Chair &
        All 
    \\
    \midrule
         SIF~\cite{genova2019learning} & 3D shapes & 
         53.0 & 65.7 & 38.9 & 49.9
         \\
         OccNet~\cite{mescheder2019occupancy} & 3D shapes & 
         57.1 & 73.7 & 50.1 & 57.1
         \\
         CvxNet~\cite{deng2020cvxnet} & 3D shapes & 
         59.8 & 67.5 & 49.1 & 56.7
         \\
         HPD~\cite{paschalidou2020learning} & 3D shapes & 
         52.9 & 70.2 & 52.6 & 58.0
    \\
    \midrule
         SoftRas~\cite{liu2019soft} & Single-view & 
         52.2 & 65.7 & 40.4 & 46.9
         \\
         VPL~\cite{kato2019learning} & Single-view & 
         53.1 & 70.1 & 45.4 & 51.3
         \\
         \ours~(ours) & Single-view & 
         \bf{57.3} & \bf{75.2} & \bf{45.5} & \bf{52.9}
    \\
    \bottomrule
    \end{tabular}
\label{tab:comparison}
\vspace{-1mm}
\end{table}

\vspace{1mm}
\noindent \textbf{Comparisons with the State-of-the-arts.}
Table~\ref{tab:comparison} shows performance comparisons against the state-of-the-art methods.
With the single-view training setting, our method performs favorably against the existing approaches.
When compared to the 3D-supervised methods, our model achieves competitive results on many object classes even though we use weaker single-view supervision.
Among the evaluated methods, SIF~\cite{genova2019learning} and CvxNet~\cite{deng2020cvxnet} can subdivide an object into fine-grained regions, and HPD~\cite{paschalidou2020learning} performs hierarchical part reasoning.
However, their shape representations require stronger supervision from 3D ground truths to produce faithful part shapes.
%
%However, their 3D representations limit the shape reconstruction and thus can not fit the objects faithfully even with stronger 3D supervision.
%
%\VJ{I am not sure if we want to say previous sentence. We either have to illustrate or show numerically why their 3D representations are limited. According to numbers, CVxNet, HPD and OccNet have much better number compared to ours.}
%
The qualitative results in Figure~\ref{fig:shapenet} demonstrate that our model discovers more faithful and consistent parts compared to the baseline methods.

\vspace{1mm}
\noindent \textbf{Ablations on Part Representations.}
We further compare 3D reconstruction methods that use different part shape constraints.
A majority of existing methods represent objects with a single mesh~\cite{kato2018neural, wang2018pixel2mesh, gkioxari2019mesh, kato2019learning, liu2019soft, henderson2020leveraging} and allow each shape to be fully deformable by predicting the vertex deformations directly. 
On the other end of the spectrum, most part-reasoning approaches represent parts with primitive shapes like cuboids~\cite{tulsiani2017learning} or super-quadratic surfaces~\cite{paschalidou2020learning}.
Our method lies within these two extremes and enables variable degree of freedom by adjusting the latent dimension of the part shape embedding.
Table~\ref{tab:shapenet_metrics} shows the quantitative comparisons between part representations like free-form meshes, cuboid reconstructions, and our Part-VAE embedding via different evaluation metrics.
In addition to 3D voxel IoU, we calculate the 2D re-projection IoU, 2D structural similarity (SSIM), and Chamfer distance (CD) between the point sets sampled from 3D volumes.
%
% Using the same 3-part setting, our model with $d=32$ produces the highest reconstruction IoU.
%
The results show that \ours~ achieves a better trade-off between the degree of deformation and shape regularization among the part-based methods.
%
% The regularization might not be sufficient with a large $d$, while a small $d$ would result in worse reconstructions due to insufficient capacity.
%
To observe how our method adapts to different object classes, we perform single-class training with different number of parts $k$ on airplane, car, and chair images.
As shown in Table~\ref{tab:single-class}, the optimal number of parts $k$ varies across object classes.
This suggests that each class have a distinct underlying part configuration to optimally represent the object shapes.
Note that \ours~achieves higher accuracy than other methods on all three classes with more than one part.
Our main reconstruction model is class-agnostic and we use the same number of parts for all classes, and yet the performance could be further improved if it is optimized for each class separately.
% Note that our method achieves greater performance gain compared to all-class training since the models can adapt to a class-specific part configuration in the single-class setting.
%
% We show the qualitative results of the ablations are also shown in Figure~\ref{fig:ShapeNet}.

\begin{table}[!t]
\centering
\caption{\textbf{Quantitative evaluations on the ShapeNet dataset~\cite{chang2015shapenet} using different metrics.}
\ours~allows part reasoning and achieves higher accuracy in terms of all metrics.}
\vspace{-2mm}
\small
\setlength\tabcolsep{4pt}
\begin{tabular}{lccccc}
\toprule
    Method & Part & 2D IoU $\uparrow$ & SSIM $\uparrow$ & 
    CD $\downarrow$ & Voxel IoU $\uparrow$
\\
\midrule
    SoftRas~\cite{liu2019soft} & & 80.5 & 86.7 & 4.76 & 46.9
\\
    VPL~\cite{kato2019learning} & & 81.0 & 88.5 & 2.65 & 51.3
\\
    Free-form & $\checkmark$ & 81.1 & 87.9 & 3.83 & 48.1
\\
    Cuboids & $\checkmark$ & 72.5 & 67.3 & 6.12 & 39.7
\\
    \ours~(ours) & $\checkmark$ & \bf{83.6 }& \bf{91.0} & \bf{2.37} & \bf{52.9}
\\
\bottomrule
\end{tabular}
\label{tab:shapenet_metrics}
\vspace{-1mm}
\end{table}

\begin{table}[!t]
\setlength\tabcolsep{10pt}
\small
    \caption{\textbf{Voxel IoU results with different number of parts $k$.}
    Note that the models are trained and tested on a single class.}
    \vspace{-2mm}
    \centering
        \begin{tabular}{lccccc}
        \toprule
            Method & $k$ & Airplane & Car & Chair \\
        \midrule
            SoftRas~\cite{liu2019soft} & 1 & 54.1 & 69.5 & 43.1 \\
            VPL~\cite{kato2019learning} & 1 & 54.6 & 74.1 & 45.3 \\
            \ours~(ours) & 1 & 54.5 & 74.3 & 45.0 \\
            \ours~(ours) & 2 & 55.4 & \bf{76.1} & 45.9 \\
            \ours~(ours) & 3 & 55.6 & 75.5 & \bf{46.6} \\
            \ours~(ours) & 6 & \bf{55.9} & 75.2 & 46.4 \\
        \bottomrule
        \end{tabular}
\label{tab:single-class}
\vspace{-1mm}
\end{table}

% ----------------------------- PartNet -----------------------------
% \vspace{1mm}
% \noindent \textbf{Evaluations on PartNet.}
\vspace{-1mm}
\subsection{Results on PartNet}
\vspace{-1mm}
To evaluate the quality of the discovered parts, we compare our results with the labeled parts in PartNet dataset.
The dataset contains hierarchical part annotations of several ShapeNet models.
We collect 111 chair samples that are in both the ShapeNet and PartNet testing sets, then combine the annotated part models into chair back, seat, and base at the coarsest level.
Note that we do not use any 3D part supervision for training, so the PartNet annotations are not ground-truths but a reference.
Since the discovered parts are not semantically labeled, we manually associate our parts to the closest corresponding PartNet annotations.
We report the quantitative voxel IoU in Table~\ref{tab:partnet} and the qualitative results in Figure~\ref{fig:partnet}.
Compared to the baseline without part prior and other representations, our method discovers more faithful parts and achieves considerably higher IoU with respect to PartNet annotations.

\begin{table}[!t]
\setlength\tabcolsep{10pt}
\small
\caption{\textbf{Voxel IoU results on the PartNet~\cite{mo2019partnet} chair samples.}}
\vspace{-2mm}
\centering
\begin{tabular}{lcccc}
    \toprule
    Method & Back & Seat & Base & Avg\\
    \midrule
    Free-form & 16.8 & 19.5 & 10.3 & 15.5\\
    Cuboids & 22.3 & 23.4 & 10.7 & 18.8\\
    \ours~(ours) & \bf{30.4} & \bf{46.0} & \bf{16.2} & \bf{30.9}\\
    \bottomrule
\vspace{-5mm}
\label{tab:partnet}
\end{tabular}
\end{table}

\begin{figure}[!t]
\centering
\setlength\tabcolsep{0pt}
\renewcommand{\arraystretch}{0}
\begin{tabular}[t]{cccc}
\vspace{-2mm}
\includegraphics[width = .20\linewidth]{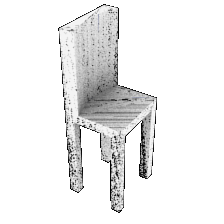} & 
\includegraphics[width = .26\linewidth]{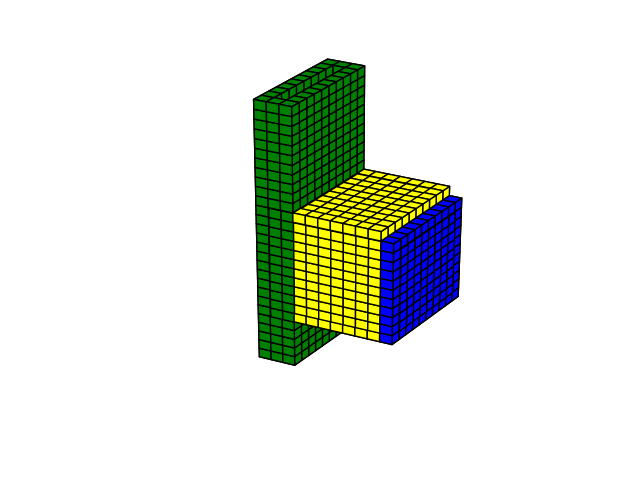} & \includegraphics[width = .27\linewidth]{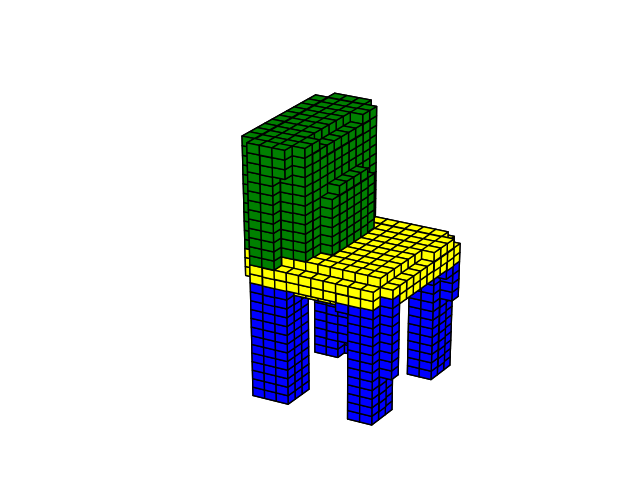} &
\includegraphics[width = .27\linewidth]{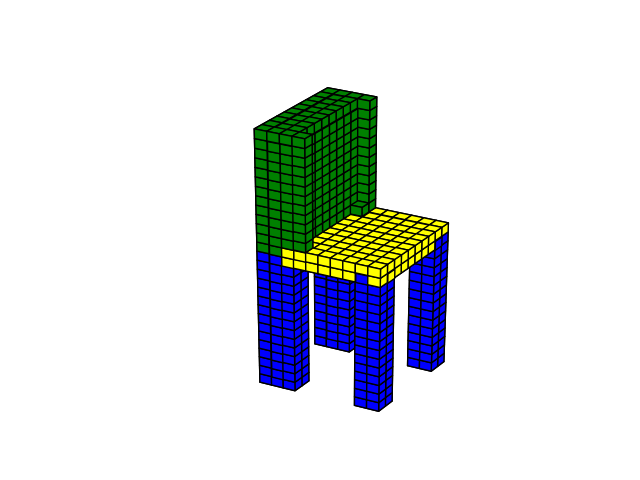}
\\
\vspace{-2mm}
\includegraphics[width = .20\linewidth]{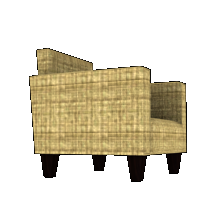} & 
\includegraphics[width = .26\linewidth]{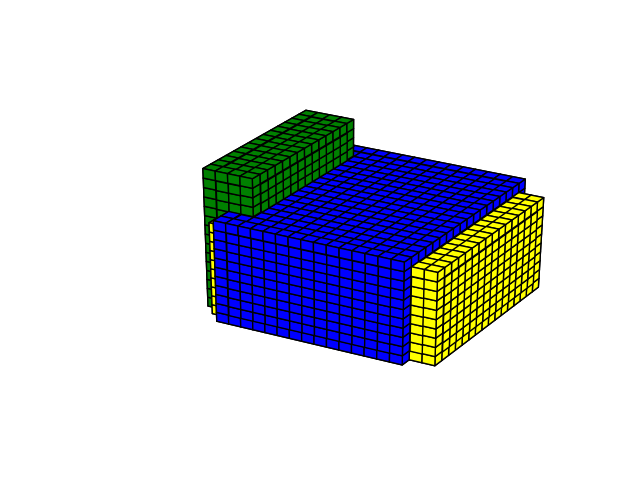} & \includegraphics[width = .27\linewidth]{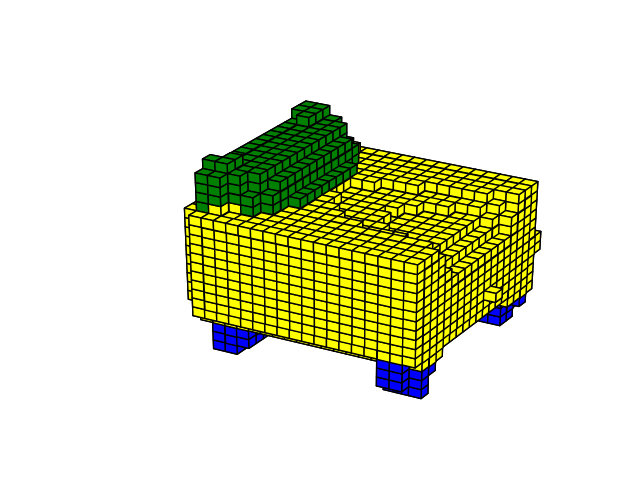} &
\includegraphics[width = .27\linewidth]{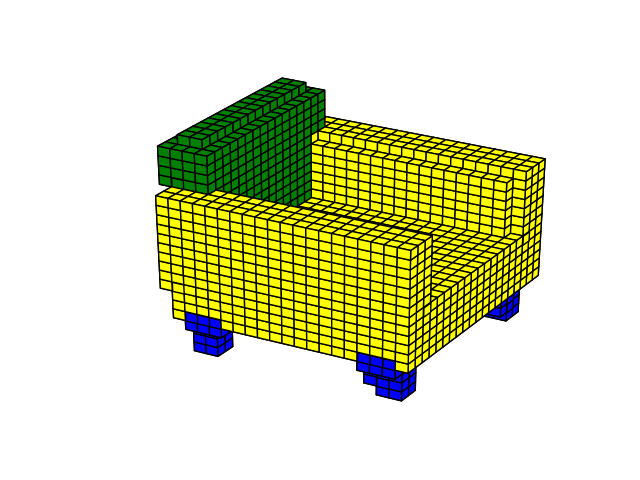}
\\
\includegraphics[width = .20\linewidth]{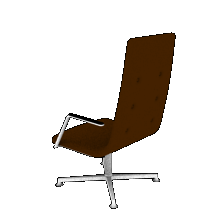} & 
\includegraphics[width = .26\linewidth]{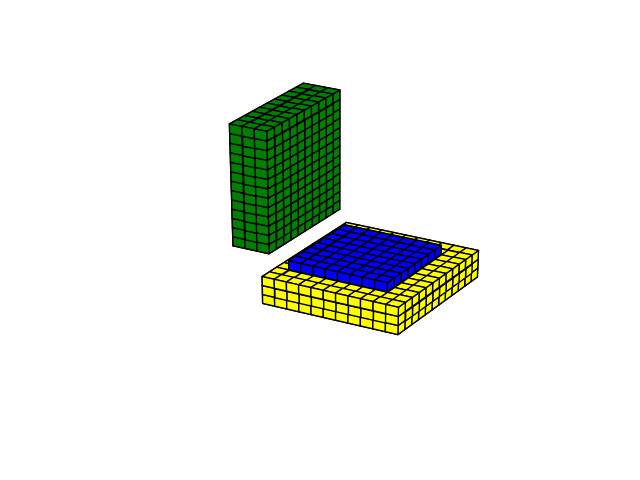} & \includegraphics[width = .27\linewidth]{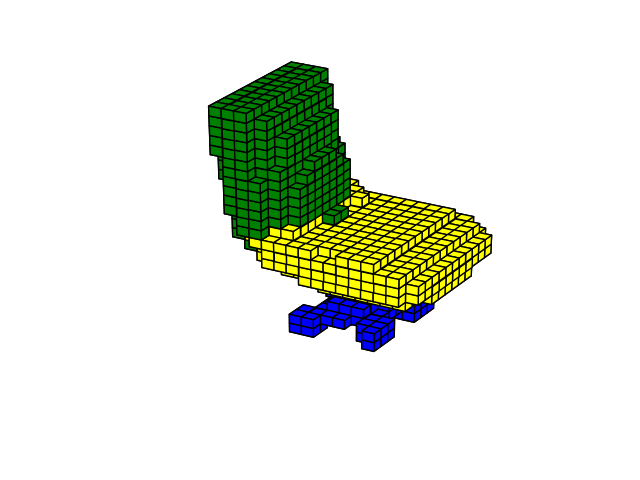} &
\includegraphics[width = .27\linewidth]{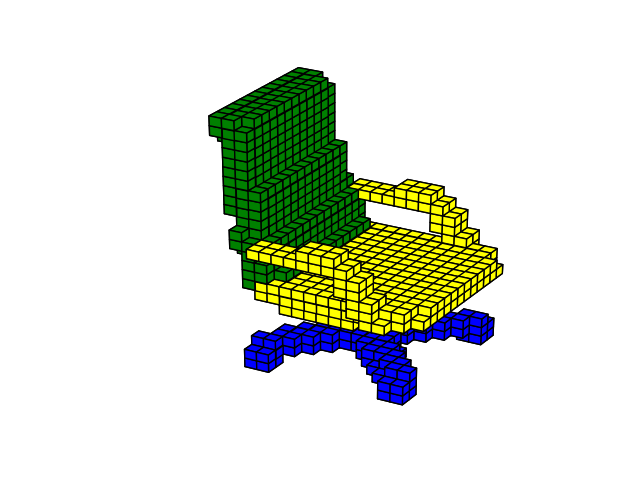}
\\
Input & 
Cuboids & 
\ours~(ours) & 
Pseudo-GT
\end{tabular}
\vspace{-1mm}
\caption{
\textbf{Qualitative results on the PartNet dataset~\cite{mo2019partnet}.}
We show the voxelized 3-part results of our method and a cuboid baseline.
Each part is specified with a color: chair back$\rightarrow$green, seat$\rightarrow$yellow, base$\rightarrow$blue.
Our method discovers faithful and consistent parts from diverse objects that
are relatively closer to the pseudo-GT part annotations in PartNet.
}
\label{fig:partnet}
\vspace{-1mm}
\end{figure}

\vspace{-1mm}
\subsection{Part Interpolation and Generation.}
\begin{figure}[!t]
\setlength\tabcolsep{0pt}
\small
\renewcommand{\arraystretch}{0}
    \begin{tabular}{ccccc}
    \vspace{-7mm}
    \includegraphics[width = .20\linewidth]{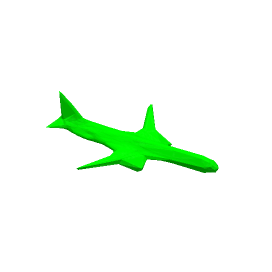} &
    \includegraphics[width = .20\linewidth]{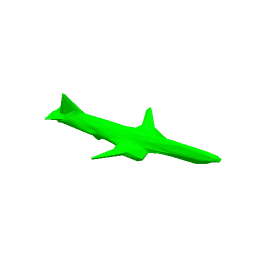} &
    \includegraphics[width = .20\linewidth]{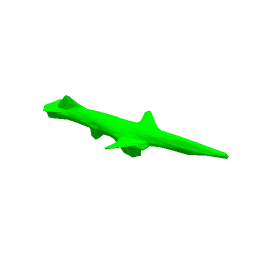} &
    \includegraphics[width = .20\linewidth]{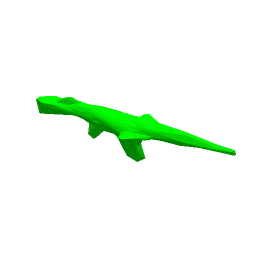} & 
    \includegraphics[width = .20\linewidth]{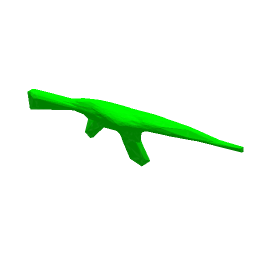}
    \\
    \vspace{-3mm}
    \includegraphics[width = .20\linewidth]{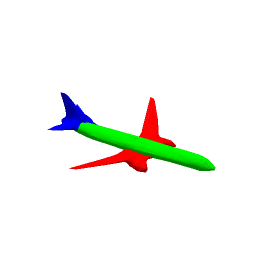} &
    \includegraphics[width = .20\linewidth]{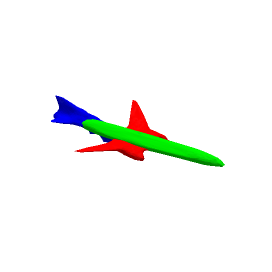} &
    \includegraphics[width = .20\linewidth]{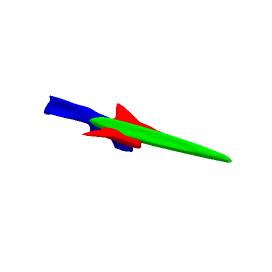} &
    \includegraphics[width = .20\linewidth]{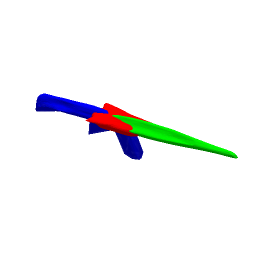} & 
    \includegraphics[width = .20\linewidth]{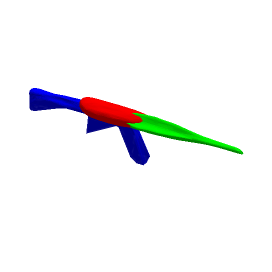}
    \\
    \includegraphics[width = .20\linewidth]{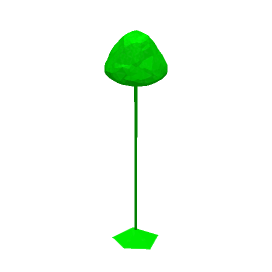} &
    \includegraphics[width = .20\linewidth]{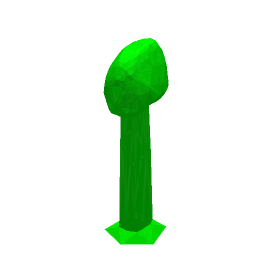} &
    \includegraphics[width = .20\linewidth]{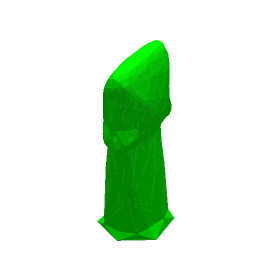} &
    \includegraphics[width = .20\linewidth]{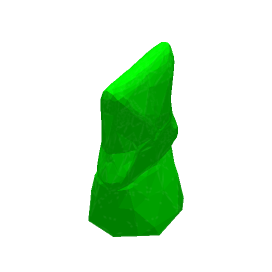} & 
    \includegraphics[width = .20\linewidth]{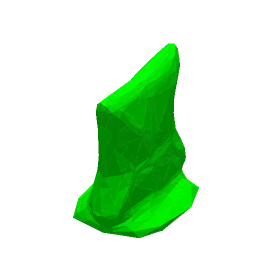}
    \\
    \includegraphics[width = .20\linewidth]{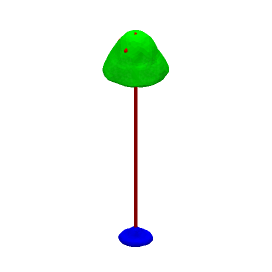} &
    \includegraphics[width = .20\linewidth]{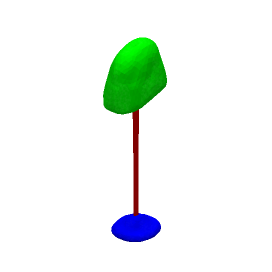} &
    \includegraphics[width = .20\linewidth]{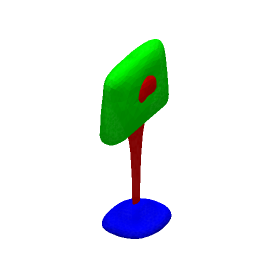} &
    \includegraphics[width = .20\linewidth]{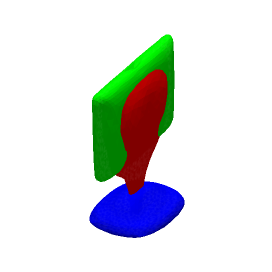} & 
    \includegraphics[width = .20\linewidth]{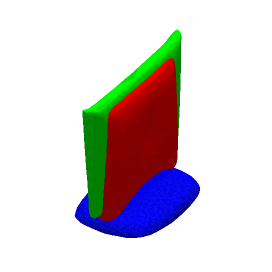}
    \\
    \includegraphics[width = .20\linewidth]{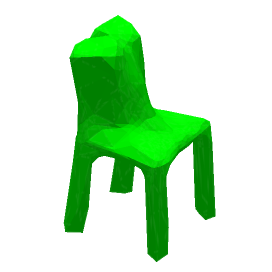} &
    \includegraphics[width = .20\linewidth]{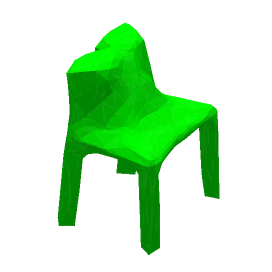} &
    \includegraphics[width = .20\linewidth]{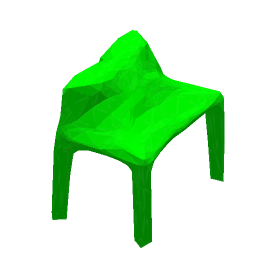} &
    \includegraphics[width = .20\linewidth]{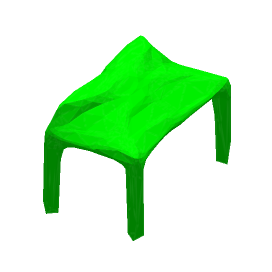} & 
    \includegraphics[width = .20\linewidth]{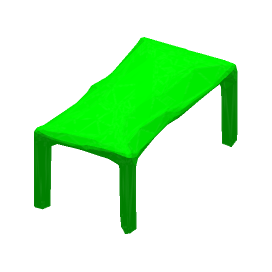}
    \\
    \includegraphics[width = .20\linewidth]{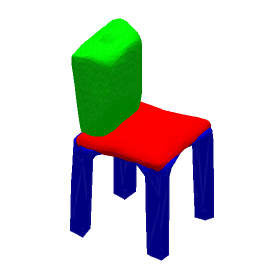} &
    \includegraphics[width = .20\linewidth]{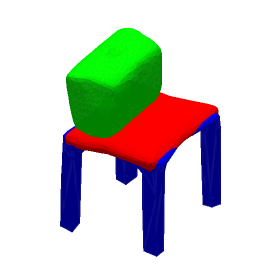} &
    \includegraphics[width = .20\linewidth]{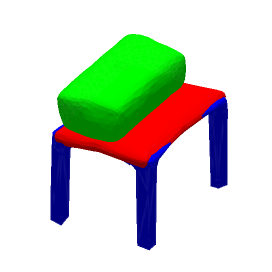} &
    \includegraphics[width = .20\linewidth]{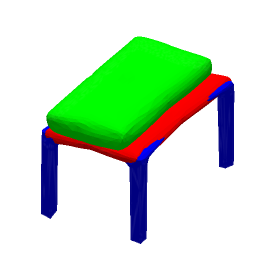} & 
    \includegraphics[width = .20\linewidth]{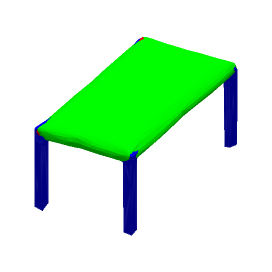}
    \\
    $\lambda$ = 0.0 & 
    $\lambda$ = 0.25 & 
    $\lambda$ = 0.5 & 
    $\lambda$ = 0.75 & 
    $\lambda$ = 1.0
    \end{tabular}
    \vspace{-1mm}
    \caption{\textbf{Cross-category interpolation.}
    We perform interpolation on ShapeNet airplane-rifle, lamp-display, and chair-table.
    We show the VPL~\cite{kato2019learning} results (mesh interpolation) in rows 1, 3, 5 and \ours~results (latent interpolation) in rows 2, 4, 6.}
\label{fig:int}
\vspace{-1mm}
\end{figure}

\begin{figure}[!t]
\setlength\tabcolsep{0pt}
\renewcommand{\arraystretch}{0}
\centering
    \begin{tabular}[t]{cc|cc}
    \includegraphics[width = .23\linewidth]{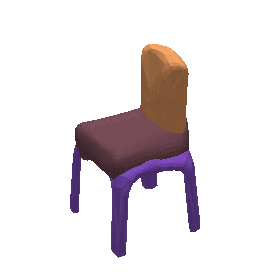} &
    \includegraphics[width = .23\linewidth]{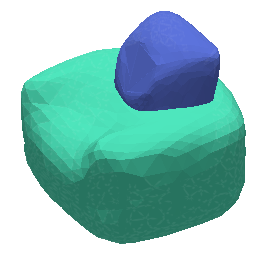} &
    \includegraphics[width = .23\linewidth]{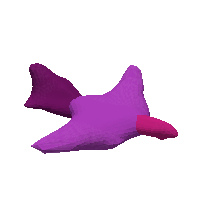} &
    \includegraphics[width = .23\linewidth]{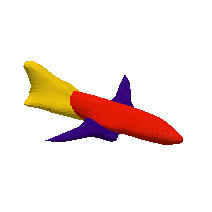}
    % \vspace{-3mm}
    \\
    \includegraphics[width = .23\linewidth]{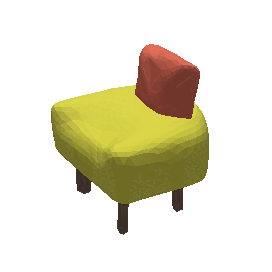} &
    \includegraphics[width = .23\linewidth]{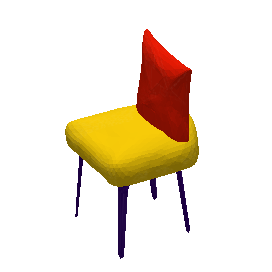} &
    \includegraphics[width = .23\linewidth]{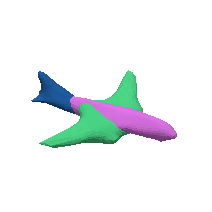} &
    \includegraphics[width = .23\linewidth]{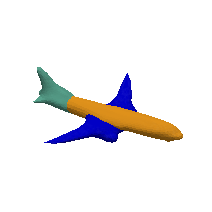}
    \end{tabular}
    \vspace{-1mm}
    \caption{\textbf{Random shape generation of chairs and airplanes.}
    We fit a GMM model on the latent shape vectors and generate random parts by sampling from individual GMM components.
    }
\label{fig:generation}
\vspace{-1mm}
\end{figure}

\vspace{-1mm}
In addition to shape reconstruction, we demonstrate two applications of Part-VAE: part-interpolation and random shape generation.
Since the object parts discovered by our models are consistent across instances, one can swap or interpolate parts to create new 3D objects.
In Figure~\ref{fig:int}, we linearly interpolate the latent encodings $(u_1, u_2)$ of two objects from different categories as: $u = \lambda u_1 + (1 - \lambda) u_2$.
Compared to the baseline without part prior, our method deforms each object part smoothly and results in more realistic shapes.
%
% Here we choose a corresponding part belonging to different object instances and linearly interpolate their latent embedding as: $u = \lambda u_1 + (1 - \lambda) u_2$, 
%
% where $u_1, u_2$ are the latent vectors of a selected part from the first and second object respectively.
%
% As demonstrated in Figure~\ref{fig:interpolation}, the selected part deforms smoothly while the other two parts remain fixed.
%
The Part-VAE can also be used as a generative model to create novel shapes.
Specifically, we fit a Gaussian Mixture Model (GMM) with $k$ components on the latent shape vectors of class-specific images.
By sampling random vectors from individual GMM distribution, we can generate $k$ random parts and combine them into a new 3D shape.
In Figure~\ref{fig:generation}, we show some randomly generated shapes of chairs and airplanes using the Part-VAE trained on the ShapeNet dataset.

% --------------------------- Pascal 3D+ -----------------------------
\vspace{-1mm}
\subsection{Results on Pascal 3D+}
\begin{figure}[!t]
\setlength\tabcolsep{0pt}
\small
\renewcommand{\arraystretch}{0}
    \begin{tabular}{ccccc}
    \includegraphics[width = .20\linewidth]{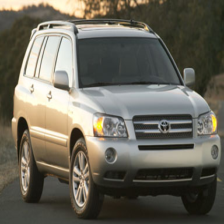} &
    \includegraphics[width = .20\linewidth]{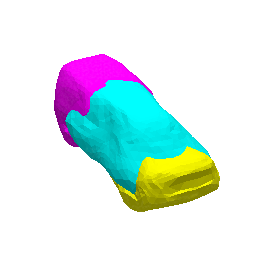} &
    \includegraphics[width = .20\linewidth]{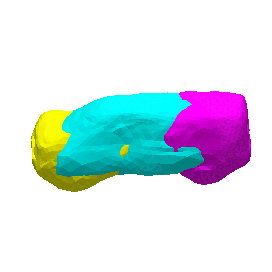} &
    \includegraphics[width = .20\linewidth]{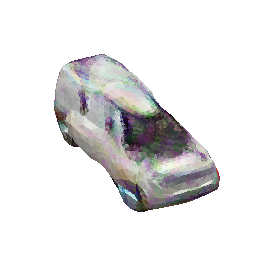} & 
    \includegraphics[width = .20\linewidth]{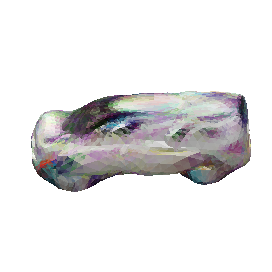}
    \\
    \includegraphics[width = .20\linewidth]{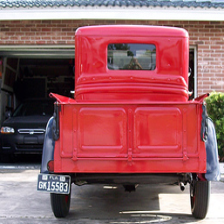} &
    \includegraphics[width = .20\linewidth]{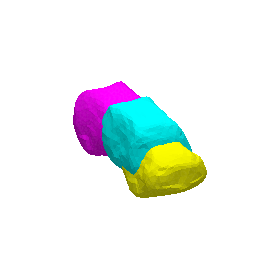} &
    \includegraphics[width = .20\linewidth]{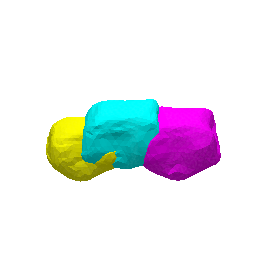} &
    \includegraphics[width = .20\linewidth]{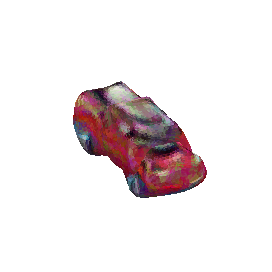} & 
    \includegraphics[width = .20\linewidth]{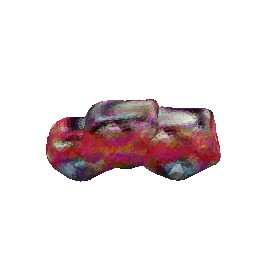}
    \\
    \includegraphics[width = .20\linewidth]{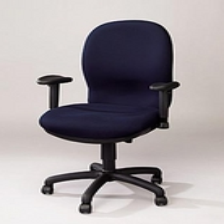} &
    \includegraphics[width = .20\linewidth]{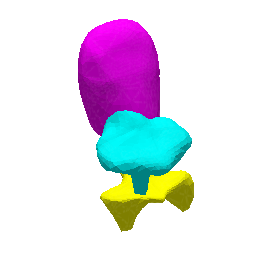} &
    \includegraphics[width = .20\linewidth]{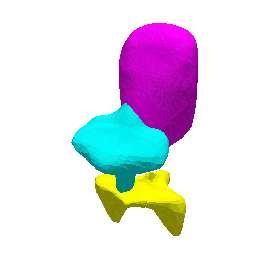} &
    \includegraphics[width = .20\linewidth]{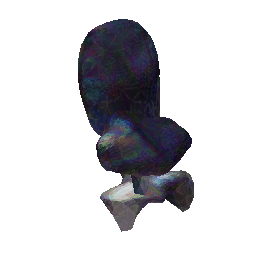} & 
    \includegraphics[width = .20\linewidth]{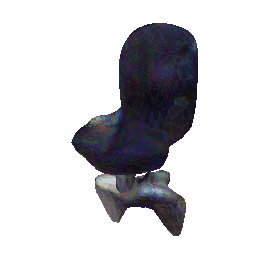}
    \\
    \includegraphics[width = .20\linewidth]{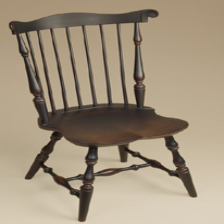} &
    \includegraphics[width = .20\linewidth]{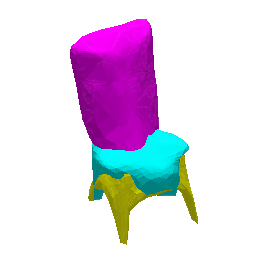} &
    \includegraphics[width = .20\linewidth]{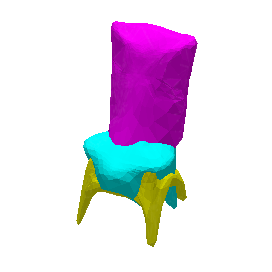} &
    \includegraphics[width = .20\linewidth]{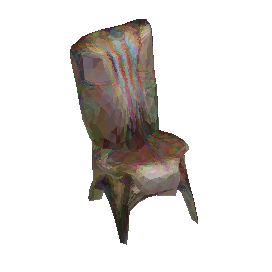} & 
    \includegraphics[width = .20\linewidth]{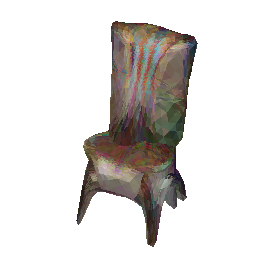}
    \\
    \includegraphics[width = .20\linewidth]{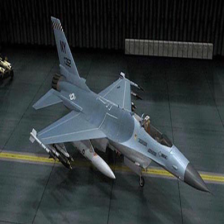} &
    \includegraphics[width = .20\linewidth]{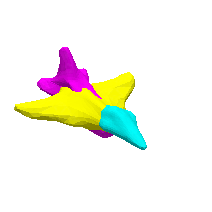} &
    \includegraphics[width = .20\linewidth]{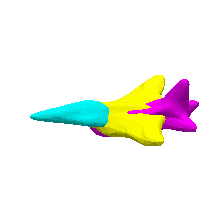} &
    \includegraphics[width = .20\linewidth]{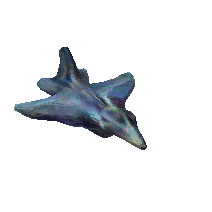} & 
    \includegraphics[width = .20\linewidth]{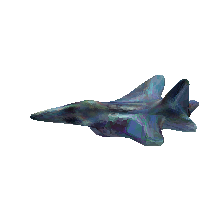}
    \\
    \includegraphics[width = .20\linewidth]{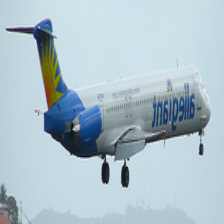} &
    \includegraphics[width = .20\linewidth]{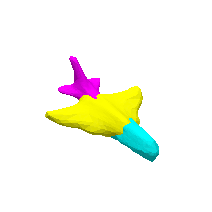} &
    \includegraphics[width = .20\linewidth]{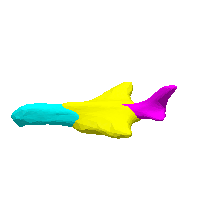} &
    \includegraphics[width = .20\linewidth]{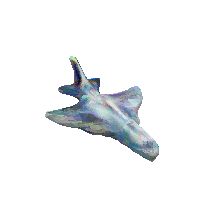} & 
    \includegraphics[width = .20\linewidth]{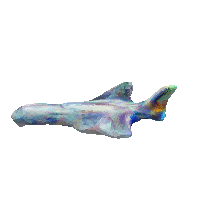}
    \\
    Input & 
    View 1 & 
    View 2 & 
    View 1 & 
    View 2
    \end{tabular}
    \vspace{-1mm}
    \caption{\textbf{Part and color reconstruction results on the Pascal 3D+ dataset~\cite{xiang2014beyond}.}
    Despite that the dataset contains complicated 3D objects in a realistic scene, our method is able to discover consistent parts and effectively reconstruct the objects shapes.
    }
\label{fig:pascal}
\vspace{-1mm}
\end{figure}

\vspace{-1mm}
We also evaluate the proposed method on the real-world images from the Pascal 3D+ dataset~\cite{xiang2014beyond} processed by Tulsiani~\etal~\cite{tulsiani2017multi}.
It consists of images in Pascal VOC~\cite{everingham2010pascal}, annotations of 3D models, silhouettes, and viewpoints in Pascal 3D+~\cite{xiang2014beyond}, and additional images in ImageNet~\cite{russakovsky2015imagenet} with silhouettes and viewpoints automatically annotated by~\cite{li2016iterative}.
This dataset is more challenging due to the complicated object shapes, image background, occlusions, and noisy silhouette annotations.
We train and evaluate our models with image resolution of 224$\times$224.
The quantitative and qualitative results are shown in Table~\ref{tab:pascal} and Figure~\ref{fig:pascal}, respectively.
Despite the challenges, our method discovers consistent object parts and achieves higher reconstruction accuracy than the state-of-the-art approaches.

\begin{table}[!t]
\small
    \caption{\textbf{Voxel IoU results on the Pascal 3D+ dataset~\cite{xiang2014beyond}.}}
    \vspace{-2mm}
    \centering
        \begin{tabular}{lccccc}
        \toprule
             Method & Part & Aeroplane & Car & Chair & Avg \\
        \midrule
             SoftRas~\cite{liu2019soft} & & 
             46.4 & 67.6 & 29.1 & 47.7 \\
             VPL~\cite{kato2019learning} & & 
             47.5 & 67.9 & 30.4 & 48.6 \\
             Free-form & \checkmark & 
             47.0 & 68.5 & 28.7 & 48.0 \\
             Cuboids & \checkmark & 
             37.1 & 60.7 & 18.9 & 38.9 \\
             \ours~(ours)  & \checkmark & 
             \bf{48.2} & \bf{69.1} & \bf{31.0} & \bf{49.4} \\
        \bottomrule
\label{tab:pascal}
\vspace{-1mm}
\end{tabular}
\end{table}

\vspace{-1mm}
\section{Concluding Remarks}
\vspace{-1mm}
In this work, we propose \ours~to discover 3D parts from single-view image collections.
By learning a part prior with Part-VAE, we demonstrate that each part can be deformed to fit a realistic object shape while constrained to have simple geometry.
With the goal to compose an object with simple parts, our reconstruction model automatically learns a latent part configuration.
In turn, the discovered parts can alleviate shape ambiguity and improve the quality of full object reconstruction.
Extensive experimental results show that \ours~can discover faithful parts from diverse object classes, and the parts are consistent across different instances within a same category.
Furthermore, we achieve the state-of-the-art reconstruction accuracy in a single-view training setting.
Our work opens up the possibilities to learn, infer, and manipulate object parts without the need for any ground-truth part labels or 3D shape supervision.

{\small
\bibliographystyle{ieee_fullname}
\bibliography{egbib}
}

\end{document}